\documentclass{article}

\usepackage{soul}
\usepackage{graphicx}
\usepackage[linesnumbered]{algorithm2e}
\usepackage{todonotes}
\usepackage{url}
\usepackage{booktabs}
\usepackage{amsfonts}
\usepackage{caption}
\usepackage{subcaption}
\usepackage{mathbbol}
\usepackage{multirow}
\usepackage{graphicx, array, blindtext}
\usepackage{amsmath}
\usepackage[capitalize,noabbrev]{cleveref}
\usepackage{sidecap}
\usepackage{sidecap}
\usepackage{caption,subcaption}
\usepackage{cleveref}
\usepackage{pbox}
\usepackage{arydshln}
\usepackage[numbers]{natbib}

\DeclareMathOperator*{\argmax}{arg\,max}

\def\defterm#1{\textbf{#1}}
\def\bs#1{\boldsymbol{#1}}
\def\set#1{\bs{#1}}

\newcommand{\numnodes}{N}
\newcommand{\unode}{i}
\newcommand{\vnode}{j}

\newcommand{\xinstance}{\set{X}}
\newcommand{\Xvariable}{\set{X}}
\newcommand{\Yvariable}{Y}

\newcommand{\gprob}{p}

\newcommand{\linstance}{\set{z}}

\newcommand{\ndim}{d} 

\newcommand{\parameter}{\theta}

\newcommand{\G}{G}
\newcommand{\A}{\set{A}} 
\newcommand{\X}{\set{X}} 
\newcommand{\Z}{\set{Z}}
 
\newcommand{\parameters}{\set{\parameter}}

\newcommand{\kld}{\mathit{KLD}}

\title{ From Graph Generation to Graph Classification}
\author{Oliver Schulte\\
School of Computing Science\\
Simon Fraser University\thanks{This research was supported by a Discovery Grant from the Natural Sciences and Engineering Research Council of Canada.}}

\begin{document}

\maketitle

\begin{abstract}
This note describes a new approach to classifying graphs that leverages graph generative models (GGM). Assuming a GGM that defines a joint probability distribution over graphs and their class labels, I derive classification formulas for the probability of a class label given a graph. A new conditional ELBO can be used to train a generative graph auto-encoder model for discrimination. While leveraging generative models for classification has been well explored for non-relational i.i.d. data, to our knowledge it is a novel approach to graph classification. 
\end{abstract}

\section{Introduction: Graph Generation and Graph Classification} \label{Introduction}

The graph classification task is to assign a discrete class label to an input graph. The dominant approach for neural graph classification is to compute an embedding for the input graph and perform the final classification in embedding space. The successful graph coarsening approach aggregates graph structural information at successively lower resolutions until a final embedding is obtained. 

Another direction for graph learning, so far unrelated, is graph generation. A graph generative model (GGM) aims to generate realistic graphs, often by sampling from a distribution over graphs. GGMs include the graph Variational Auto-Encoder (GVAE), auto-regressive methods, and most recently graph diffusion models. 

For non-relational i.i.d. data, it is well-known that a classification model can be derived from a generative model~\cite[Ch.1.5.4]{Bishop2006}. The advantages and disadvantages of a generative approach to classification have been extensively researched, both theoretically and empirically. To my knowledge, this is the first note to study the generative approach for graph classification. For non-relational data, previous work has found two important advantages for the generative approach.

\begin{enumerate}
    \item Leveraging insights from sophisticated generative models.
    \item Faster convergence to optimal classification for smaller sample sizes.
\end{enumerate}

Both points are potentially valuable for graph classification. Because graph data are much more complex than i.i.d. data, much work has gone into developing sophisticated GGMs. It is desirable to leverage this work for graph classification. It is also common for graph data sets to contain a fairly small number of graphs (while each graph may be large); in fact, often graph data come in the form of just a single network. 

This note derives formulas for classifying graphs based on a class-conditional GGM. The class-conditional GGMs can be trained with a logistic objective. For  generative models based on VAEs, a new logistic discriminative training objective is given, based on a variant of a conditional ELBO. 

{\em Please direct comments, suggestions, and proposals for collaboration to \tt{oschulte@cs.sfu.ca}.}

\section{Logistic Graph Classification Model} \label{sec:classification}

An attributed graph is a pair $\G=(V,E)$ comprising a finite set of $\numnodes$ nodes and edges where each node  is assigned an $\ndim$-dimensional attribute $\xinstance_{\unode}$. An attributed graph can be represented by an $\numnodes\times \numnodes$ adjacency matrix $\A$ with $\{0,1\}$ entries, together with an $\numnodes \times \ndim$ node feature matrix $\X$. 

We want to build a graph classifier that maximizes the cross-entropy objective:

\begin{align} \label{eq:cross-entropy}
\sum_{j=1}^{m} y^{j}\big(\ln P(y=1|\A,\Xvariable) + 
      (1- y^{j})  \big(\ln (1-P(y=1|\A,\Xvariable))\big)
\end{align}

We assume a class-conditional GGM that defines a joint distribution over graphs given class labels $P(\A,\X|\Yvariable)$. It is well-known that a classifier can be derived from a class-conditional GGM through the log-odds~\cite[Ch.4.2]{Bishop2006}:

\begin{align} \label{eq:gen2cond} 
P(y=1|\A,\Xvariable) = \sigma(L(\A,\Xvariable))\\
L(\A,\Xvariable) =_{df}
   \ln{\frac{P(y=1|\A,\Xvariable)}{P(y=0|\A,\Xvariable)}} =  \ln{\frac{P(\A,\Xvariable|y=1)}{P(\A,\Xvariable|y=0)}} + \ln{\frac{P(y=1)}{P(y=0)}} \notag
\end{align} 

The cross-entropy objective~\Cref{eq:cross-entropy} for the logistic classifier~\Cref{eq:gen2cond} is given by

\begin{align} \notag
\argmax_{\set{\parameters}}
\sum_{j=1}^{m} y^{j}\big(\ln \sigma(L_{\set{\parameters}}(\A^j,\Xvariable^j)) + 
      (1- y^{j})  \big(\ln (1-\sigma(L_{\set{\parameters}}(\A^j,\Xvariable^j))\big)
\end{align}
where $\set{\parameters}$ are the parameters of the class-conditional GGM $P_{\set{\parameters}}(\A,\X|\Yvariable)$.





\section{Logistic Variational Auto-Encoder}

In principle any class-conditional GGM can be used, such as GraphVAEs, auto-regressive, and diffusion models. In this section we consider an approach based on a general variational auto-encoder, which could be for example a GraphVAE~\cite{DBLP:conf/icann/SimonovskyK18} or a VGAE~\cite{kipf2016variational}. We propose the following evidence lower bound (ELBO) objective for the logistic cross-entropy.

\begin{align} 
      \sum_{j=1}^{m} y^{j} \ln{\gprob(Y=1|\A^{j},\Xvariable^{j})} + (1-y^{j}) \ln{\gprob(Y=0|\A^{j},\Xvariable^{j})} \geq \notag \\
       \sum_{j=1}^{m} y^{j}\big(E_{z \sim q(z|\A^j,\Xvariable^j)}[\ln \sigma(L(\A^j,\Xvariable^j|\linstance))]) + 
      (1- y^{j}\big) \big(E_{z \sim q(z|\A^j,\Xvariable^j)}[\ln (1-\sigma(L(\A^j,\Xvariable^j|\linstance))]\big)\label{eq:our-objective}\\
      L(\A^j,\Xvariable^j|\linstance)) = \ln{\frac{P(y=1|\A,\Xvariable,\linstance)}{P(y=0|\A,\Xvariable,\linstance)}} =  \ln{\frac{P(\A,\Xvariable|\linstance,y=1)}{P(\A,\Xvariable|\linstance,y=0)}} + \ln{\frac{P(y=1|\linstance)}{P(y=0|\linstance)}} \notag
\end{align}

Here the latent variable $\Z$ can either range over graph embeddings, as in a GraphVAE~\cite{DBLP:conf/icann/SimonovskyK18}, or node embeddings, as in a VGAE~\cite{kipf2016variational}. The objective~\Cref{eq:our-objective} can be implemented by the following architecture.

    \textbf{Encoder.} Use any graph embedding method, for $q(\linstance|\A,\Xvariable)$, including graph embedding methods developed for graph classification~\cite{Errica2020A}. For node embeddings, any graph neural network encoder can be used. If we use graph embeddings, we can view the logistic variational approach to graph classification as adding a powerful novel graph decoder to a standard graph encoder.
    
    \textbf{Decoder.} Use a class-conditional graph generative model to compute the class log-odds $L(\A^j,\Xvariable^j|\linstance)$. For example, if $\linstance$ is a graph embedding, then $P(\A,\Xvariable|\linstance,y)$ could be implemented by an MLP that maps $\linstance$ and $y$ to an adjacency matrix, as described in ~\cite{DBLP:conf/icann/SimonovskyK18}. If $\linstance$ is a matrix of node embeddings, then the decoder model can map two node embeddings $\linstance_{\unode}$ and $\linstance_{\vnode}$ to a link probability. 
    
    \textbf{Class Prior.} A baseline implementation assumes that the class label is independent of the latent variable and  uses the observed class frequency as an estimate. In symbols, $P(y=1|\linstance) = \hat{P}(y=1)$ where $\hat{P}$ is the proportion of positive graphs observed in the training sample.


\section{Derivation of ELBO}

This section derives the objective inequality~\eqref{eq:our-objective}. 
We start with a conditional ELBO 
developed by~\citeauthor{sohn2015learning}~\citeyearpar{sohn2015learning}:

\begin{align} \label{eq:celbo}
  \ln  \gprob(Y|\A,\Xvariable) 
     \geq E_{z \sim q(z|\A,\Xvariable,Y)}[\ln \gprob(Y|A,\Xvariable, z)] - \kld(q(Z|\A,\Xvariable,Y)||p(Z|\A,\Xvariable))
\end{align}

where $p(Z|\A,\Xvariable)$ represents the \defterm{prior network} and $q(Z|\A,\Xvariable,Y=1)$ the \defterm{recognition network.} 

A potential problem with~\cref{eq:celbo} is information leakage from the encoder that allows the decoder to ``cheat". For example, suppose the encoder simply sets $z = y$. Then the decoder could set $p(\cdot|y,z) = 99\%$ if $y=z$ and $1\%$ otherwise. \citeauthor{sohn2015learning} address this by setting the recognition network equal to the prior network, resulting in a a Gaussian stochastic neural network (GSNN). 

\begin{equation} \label{eq:gsnn}
    q(Z|\A,\Xvariable,Y) = p(Z|\A,\Xvariable)
\end{equation}

\citeauthor{sohn2015learning} mix the GSNN approximation where $p = q$ with a CVAE approximation, where $p \neq q$. A simpler alternative is starting with the GSNN objective first, then consider mixing if the class-conditional generation GSNN looks promising. Combining the ELBO~\eqref{eq:celbo} with the GSNN condition~\eqref{eq:gsnn} gives the GSNN ELBO

\begin{equation} \label{eq:gsnn-elbo}
\ln  \gprob(Y|\A,\Xvariable) 
     \geq E_{z \sim p(z|\A,\Xvariable)}[\ln \gprob(Y|A,\Xvariable, z)].
\end{equation}

Using~\Cref{eq:gsnn-elbo} as an approximate classification model, a cross-entropy ELBO is given by

\begin{align} 
\sum_{j=1}^{m} y^{j} 
\ln{\gprob(Y=1|\A^{j},\Xvariable^{j})} + (1-y^{j}) \ln{\gprob(Y=0|\A^{j},\Xvariable^{j})} \geq \notag \\
 \sum_{j=1}^{m}    y^{j}\big(E_{z \sim p(z|\A^j,\Xvariable^j)}[\ln \gprob(Y=1|\A,\Xvariable,\linstance)]) + 
      (1- y^{j}\big) \big(E_{z \sim p(z|\A^j,\Xvariable^j)}[\ln (1-\gprob(Y=1|\A,\Xvariable,\linstance))]\big) = \label{eq:cross-entropy-gsnn}\\
       \sum_{j=1}^{m} y^{j}\big(E_{z \sim p(z|\A^j,\Xvariable^j)}[\ln \sigma(L(\A^j,\Xvariable^j|\linstance))]) + 
      (1- y^{j}\big) \big(E_{z \sim p(z|\A^j,\Xvariable^j)}[\ln (1-\sigma(L(\A^j,\Xvariable^j|\linstance)))]\big) \label{eq:cross-entropy-logistic}
\end{align}

~\Cref{eq:cross-entropy-logistic} follows by applying the logistic classification model~\Cref{eq:gen2cond} to~\Cref{eq:cross-entropy-gsnn}. Our proposed objective~\Cref{eq:our-objective} follows from~\Cref{eq:cross-entropy-logistic} simply by changing the notation $p(z|\A^j,\Xvariable^j)$ to $q(z|\A^j,\Xvariable^j)$. 



\bibliographystyle{plainnat}
\bibliography{main.bib,master.bib}

\begin{thebibliography}{5}
\providecommand{\natexlab}[1]{#1}
\providecommand{\url}[1]{\texttt{#1}}
\expandafter\ifx\csname urlstyle\endcsname\relax
  \providecommand{\doi}[1]{doi: #1}\else
  \providecommand{\doi}{doi: \begingroup \urlstyle{rm}\Url}\fi

\bibitem[Bishop(2006)]{Bishop2006}
Christopher~M. Bishop.
\newblock \emph{Pattern Recognition and Machine Learning}.
\newblock Springer, 2006.

\bibitem[Errica et~al.(2020)Errica, Podda, Bacciu, and Micheli]{Errica2020A}
Federico Errica, Marco Podda, Davide Bacciu, and Alessio Micheli.
\newblock A fair comparison of graph neural networks for graph classification.
\newblock In \emph{International Conference on Learning Representations}, 2020.
\newblock URL \url{https://openreview.net/forum?id=HygDF6NFPB}.

\bibitem[Kipf and Welling(2016)]{kipf2016variational}
Thomas Kipf and M.~Welling.
\newblock Variational graph auto-encoders.
\newblock \emph{ArXiv}, abs/1611.07308, 2016.

\bibitem[Simonovsky and Komodakis(2018)]{DBLP:conf/icann/SimonovskyK18}
Martin Simonovsky and Nikos Komodakis.
\newblock Graph{VAE}: Towards generation of small graphs using variational
  autoencoders.
\newblock In \emph{Artificial Neural Networks and Machine Learning}, volume
  11139 of \emph{Lecture Notes in Computer Science}, pages 412--422, 2018.

\bibitem[Sohn et~al.(2015)Sohn, Lee, and Yan]{sohn2015learning}
Kihyuk Sohn, Honglak Lee, and Xinchen Yan.
\newblock Learning structured output representation using deep conditional
  generative models.
\newblock \emph{Advances in neural information processing systems},
  28:\penalty0 3483--3491, 2015.

\end{thebibliography}

\end{document}